 \documentclass[final,5p,times,twocolumn]{elsarticle}

\usepackage{amssymb}
\usepackage{multirow}
\usepackage{booktabs}
\usepackage{wrapfig} 
\usepackage{flushend}
\usepackage{subfigure}
\journal{Pattern Recognition}

\begin{document}

\begin{frontmatter}

%% Title, authors and addresses

%% use the tnoteref command within \title for footnotes;
%% use the tnotetext command for theassociated footnote;
%% use the fnref command within \author or \address for footnotes;
%% use the fntext command for theassociated footnote;
%% use the corref command within \author for corresponding author footnotes;
%% use the cortext command for theassociated footnote;
%% use the ead command for the email address,
%% and the form \ead[url] for the home page:
%% \title{Title\tnoteref{label1}}
%% \tnotetext[label1]{}
%% \author{Name\corref{cor1}\fnref{label2}}
%% \ead{email address}
%% \ead[url]{home page}
%% \fntext[label2]{}
%% \cortext[cor1]{}
%% \affiliation{organization={},
%%             addressline={},
%%             city={},
%%             postcode={},
%%             state={},
%%             country={}}
%% \fntext[label3]{}

\title{Exploring Low-Resource Medical Image Classification with Weakly Supervised Prompt Learning}

%% use optional labels to link authors explicitly to addresses:
%% \author[label1,label2]{}
%% \affiliation[label1]{organization={},
%%             addressline={},
%%             city={},
%%             postcode={},
%%             state={},
%%             country={}}
%%
%% \affiliation[label2]{organization={},
%%             addressline={},
%%             city={},
%%             postcode={},
%%             state={},
%%             country={}}

\author[1]{Fudan Zheng\fnref{fn1}}
\fntext[fn1]{Equal contribution.}

\author[1]{Jindong Cao\fnref{fn1}}

\author[2]{Weijiang Yu\corref{cor1}}
\cortext[cor1]{Corresponding author:}
\ead{weijiangyu8@gmail.com}

\author[1]{Zhiguang Chen}

\author[1]{Nong Xiao}

\author[1]{Yutong Lu\corref{cor1}}
\ead{yutong.lu@nscc-gz.cn}

\affiliation[1]{organization={Sun Yat-Sen University},%Department and Organization
            addressline={No. 132 Waihuandong Road, Guangzhou Higher Education Mega Center}, 
            city={Guangzhou},
            postcode={510006}, 
            country={China}}
        
\affiliation[2]{organization={Huawei Technologies Co.,Ltd.},%Department and Organization
       	addressline={Huawei Industrial Park, Bantian, Longgang District}, 
       	city={Shenzhen},
       	postcode={518129}, 
       	country={China}}

\begin{abstract}
Most advances in medical image recognition supporting clinical auxiliary diagnosis meet challenges due to the low-resource situation in the medical field, where annotations are highly expensive and professional. This low-resource problem can be alleviated by leveraging the transferable representations of large-scale pre-trained vision-language models like CLIP. After being pre-trained using large-scale unlabeled medical images and texts (such as medical reports), the vision-language models can learn transferable representations and support flexible downstream clinical tasks such as medical image classification via relevant medical text prompts. However, existing pre-trained vision-language models require domain experts (clinicians) to carefully design the medical text prompts based on different datasets when applied to specific medical image tasks, which is extremely time-consuming and greatly increases the burden on clinicians. To address this problem, we propose a weakly supervised prompt learning method $MedPrompt$ for automatically generating medical prompts, which includes an unsupervised pre-trained vision-language model and a weakly supervised prompt learning model. The unsupervised pre-trained vision-language model adopts large-scale medical images and texts for pre-training, utilizing the natural correlation between medical images and corresponding medical texts without manual annotations. The weakly supervised prompt learning model only utilizes the classes of images in the dataset to guide the learning of the specific class vector in the prompt, while the learning of other context vectors in the prompt does not require any manual annotations for guidance. To the best of our knowledge, this is the first model to automatically generate medical prompts. With the assistance of these prompts, the pre-trained vision-language model can be freed from the strong expert dependency of manual annotation and manual prompt design, thus achieving end-to-end, low-cost medical image classification. Experimental results show that the model using our automatically generated prompts outperforms all its hand-crafted prompts counterparts in full-shot learning on all four datasets, and achieves superior accuracy on zero-shot image classification and few-shot learning in three of the four medical benchmark datasets and comparable accuracy in the remaining one. In addition, the proposed prompt generator is lightweight and therefore has the potential to be embedded into any network architecture.

\end{abstract}

%%Graphical abstract
%\begin{graphicalabstract}
%\includegraphics{grabs}
%\end{graphicalabstract}

%%Research highlights
%%\begin{highlights}
%%\item Research highlight 1
%%\item Research highlight 2
%%\end{highlights}

\begin{keyword}
	Medical image classification \sep Weakly supervised learning \sep Prompt learning \sep Few-shot learning \sep Zero-shot learning

%% keywords here, in the form: keyword \sep keyword

%% PACS codes here, in the form: \PACS code \sep code

%% MSC codes here, in the form: \MSC code \sep code
%% or \MSC[2008] code \sep code (2000 is the default)

\end{keyword}

\end{frontmatter}

\section{Introduction}

Medical imaging techniques, such as computed tomography (CT), magnetic resonance imaging (MRI), and X-rays, are often used in clinical practice for monitoring, diagnosis, and treatment. With the massive growth of medical imaging data and the rapid development of deep learning technology, researchers have developed various deep learning models applied to medical images to support clinical decision-making, and these models have been proven to have high accuracy and generalization ability \cite{1,2,3}. Existing models generally adopt the supervised learning mode, requiring medical images to have corresponding complete, exact, and accurate annotations. Unlike natural image annotation, which can be conducted by ordinary people, medical image annotation must be carried out by experienced domain experts (clinicians), with an extremely high annotation threshold and cost. This directly leads to the current situation of low-resource medical data —— there are limited available annotated medical samples, while a large number of unlabeled medical images and text (such as medical reports) samples are left behind and underutilized.

The emergence of large-scale pre-trained vision-language models such as CLIP \cite{4} makes it possible for unlabeled samples to be fully utilized. Some studies have employed such models to make the most of the enormous volume of unlabeled medical images and texts \cite{5,6,7}. These large-scale vision-language models are pre-trained by predicting the correct matching between image and text pairs, or the similarities between images and texts, to learn transferable image and text representations to support flexible downstream clinical tasks. Transferring the pre-trained models to specific downstream clinical tasks is often done with the help of relevant medical text prompts. Taking the large-scale medical pre-trained vision-language model MedCLIP \cite{7} as an example, specifically, when using the pre-trained MedCLIP model for medical image classification, the image and multiple different medical text prompts associated with different predicted categories are fed into the model's image encoder and text encoder respectively to obtain their corresponding embeddings. Then, the model calculates the similarities between the image embeddings and these text embeddings, and takes the category of the text embeddings with the highest similarity to the image embeddings as the category of the image. The quality of the input text prompts, even a slight change in wording, has been proven to have a significant impact on model performance \cite{8}. Therefore, there is a significant reliance on domain experts (clinicians) when generating medical text prompts.

\begin{figure*}[htbp]
	\centerline{\includegraphics[width=0.7\textwidth]{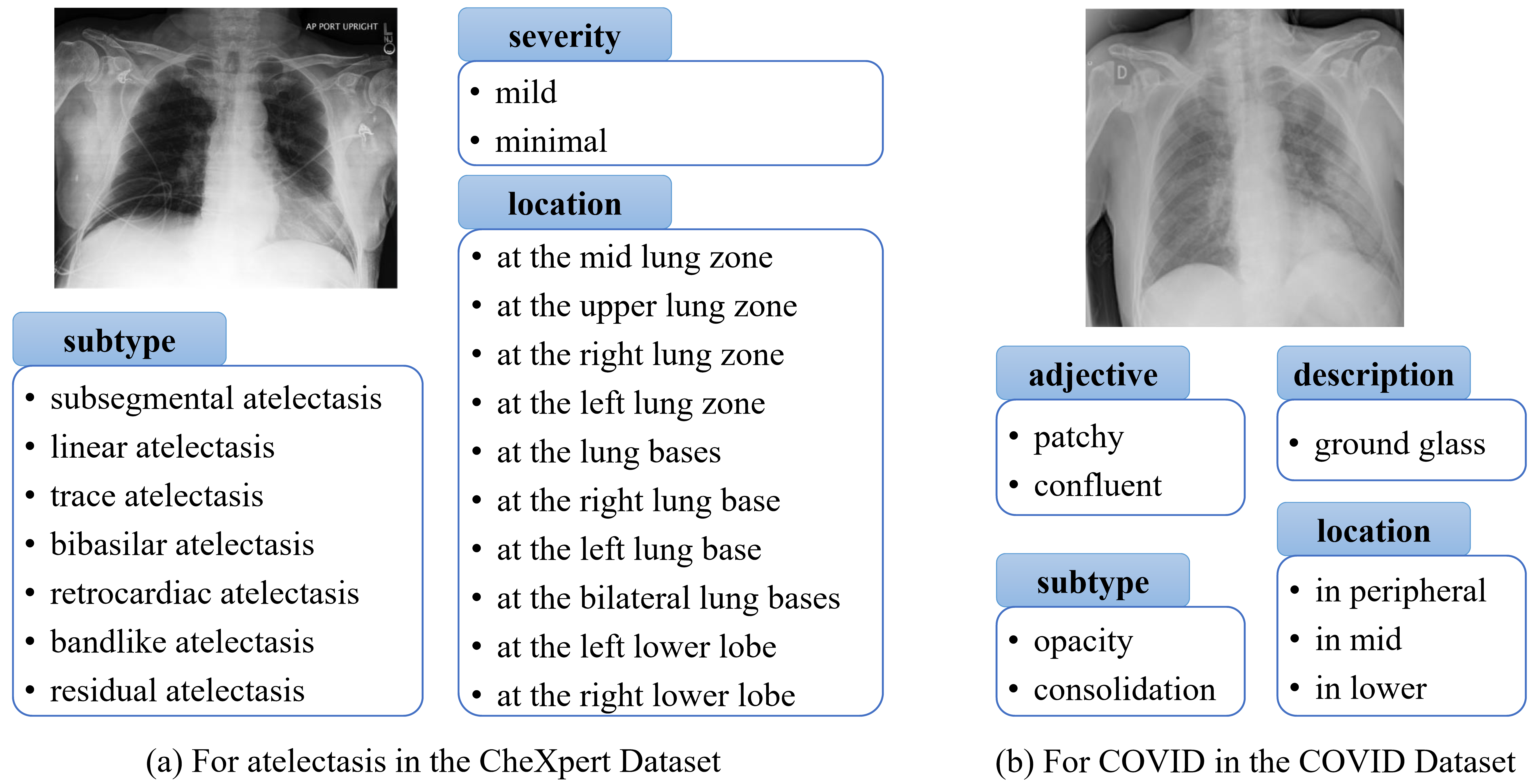}}
	\caption{Examples of prompts manually designed by clinicians for different categories in different datasets. (a) For atelectasis in the CheXpert Dataset. (b) For COVID in the COVID Dataset. It can be seen that these text prompts are highly related to the characteristics of images in the dataset and strongly dependent on domain experts and domain knowledge.}
	\label{Fig_1_manual_prompts}
\end{figure*}

Figure \ref{Fig_1_manual_prompts} shows some examples of prompts carefully designed by clinicians according to different categories in different datasets. It can be seen that the prompts designed, such as disease subtypes, severity, and location, are highly related to the characteristics of images in the dataset, have significant differences in the category and specific content, and are highly targeted and professional. Thus, manually designing text prompts is extremely time-consuming and greatly dependent on domain experts and domain knowledge, which obviously goes against the original intention of employing large-scale pre-trained vision-language models in the field of low-resource medical images to minimize the reliance on clinicians. Therefore, automatically generating prompts is a better solution.

To this end, we propose a weakly supervised prompt learning framework $MedPrompt$, which can generate high-quality medical text prompts automatically. The framework includes an unsupervised pre-trained vision-language model and a weakly supervised prompt learning model. The unsupervised pre-trained vision-language model adopts large-scale medical images and texts for pre-training, utilizing the natural correlation between medical images and corresponding medical texts without manual annotations. The weakly supervised prompt learning model (i.e. the prompt generator) consists of a learnable two-layer bottleneck structure and the projection operations of context embeddings and class embeddings. It only utilizes the classes of images in the dataset to guide the learning of the specific class vector in the prompt, while the learning of other context vectors in the prompt does not require any manual annotations for guidance. The framework can be trained to generate high-quality prompts automatically. Benefiting from these prompts, the pre-trained vision-language model utilizes the similarity between image embeddings and prompts embeddings to determine the categories of the images. Throughout the process, the framework is almost completely free from expert reliance on manual annotations and manual prompt design, thus enabling end-to-end, low-cost medical image recognition. Experimental resultsshow that: 1) On CheXpert, MIMIC-CXR and COVID, the zero-shot interference of our proposed model outperforms the zero-shot interference and even the full-shot learning of all its hand-crafted prompts counterparts; 2) On RSNA, with the extra cost of only 4 and 16 labeled samples for few-shot learning, our proposed model outperforms all its hand-crafted prompts counterparts in zero-shot interference, and is comparable to the SOTA model in full-shot learning, respectively; 3) On all four datasets, our proposed model exceeds all its hand-crafted prompts counterparts in full-shot learning. Moreover, the module used to automatically generate prompts is a lightweight module that does not impose too many additional network parameters and computational burdens on the model, which has the potential to be embedded into any network architecture, whether into a large-scale network architecture or into an end-edge or mobile network architecture. In summary, our work has the following five main contributions:

1. We propose a weakly supervised prompt learning framework $MedPrompt$, which enables the model to directly learn from the datasets and automatically generate effective prompts by using only class labels for inexact supervised learning, saving the expensive cost and intensive effort of experts in manual prompt design;

2. To the best of our knowledge, this is the first model to automatically generate medical prompts;

3. In medical image classification tasks on all four benchmark datasets, our model outperforms all its hand-crafted prompts counterparts in fully supervised learning;

4. In three of the four datasets, the zero-shot classification performance of our model exceeds the zero-shot and even the fully supervised performance of existing models, demonstrating its superior generalization ability; and on the remaining dataset, with only a tiny sample and training cost for few-shot learning, our model outperforms all its hand-crafted prompts counterparts in zero-shot interference and is comparable to the SOTA model in full-shot learning;

5. The module automatically generating prompts is lightweight and has the potential to be embedded into any network architecture.

\section{Related work}
\subsection{Medical vision-language models}
Large-scale pre-trained vision-language models have been widespread in general domains. However, due to the relatively small amount of medical image and text data available for pre-training, the pre-trained vision-language models in the medical domain are still under exploration.

Following the general pre-trained vision-language models, existing medical image-text representation learning also adopts the framework of contrast learning. Among them, most of the work used strictly paired medical images and texts for contrast learning \cite{5,6,9}, which not only reduced the number of paired images and texts available for pre-training but also introduced false negative noise during the training process. To solve these problems, Wang et al. \cite{7}  proposed to decouple the strong pairing relationship between image and text to obtain more usable training data and eliminate false negatives. Accordingly, a semantic matching loss was used to replace the commonly used InfoNCE loss in the original vision-language models. Following Wang's work, we extended the pre-trained medical image and text pairs to enable the model to be pre-trained on a larger scale of image-text data.

\subsection{Prompt learning}
Prompt learning was initially a vital research hotspot in natural language processing. The motivation is to leverage the pre-trained language models (such as BERT \cite{10} or GPT \cite{11,12}) as knowledge bases from which prompt templates can be used to elicit valuable information for downstream tasks \cite{15}. After pre-training the model using a large amount of original text, a prompt function can be designed to adapt the pre-trained model to small or unlabeled data in other scenarios for few-shot learning or zero-shot inference.

Manual template design is commonly used in previous work, and the designed templates have achieved good performance in cloze test, question answering, translation, text classification, and conditional text generation tasks \cite{17,18,19}. However, creating and verifying these prompts require time and experience; even the most experienced prompt designers may have difficulties discovering the best prompts manually \cite{20,21}. Therefore, many approaches have been proposed to generate prompts automatically. For example, Jiang et al. propose mining-based and paraphrasing-based methods to automatically generate high-quality and diverse prompts, which more accurately estimate the knowledge contained in language models \cite{20}. Shin et al. develop an automated method based on a gradient-guided search to create prompts for a diverse set of tasks \cite{22}. Other studies transform the prompt into continuous vectors and optimize the objective function end to end \cite{24,25}.

CoOp \cite{8} and CoCoOp \cite{26} are the first to apply prompt learning to the adaptation of large vision-language models in computer vision. CoCoOp is a continuous prompt learning method that can automatically construct conditional contextual prompts. It consists of a set of context prompt vectors and a lightweight neural network (Meta-Net), which generates an input-conditional token (vector) for each image. CoCoOp has achieved excellent zero-shot inference performance on previously unseen categories under low-resource conditions. Inspired by CoCoOp, our prompt generator for learning medical text prompts also learns relevant representations based on each image instance.

\subsection{Weakly-supervised learning}
In many tasks, obtaining strong supervision information is difficult due to the high cost of manual data labeling. Weakly supervised learning is proposed to be applied to such low-resource scenarios that lack enough correct manual labeling, aiming at building prediction models through weak supervision signals. There are three typical types of weak supervision: incomplete supervision, inexact supervision, and inaccurate supervision \cite{27}. Among them, inexact supervision means that the training data has only coarse-grained labels but no exact labels corresponding to the training target. In the prompt learning process of this work, we also do not have exact labels (exact natural language or semantic labels) for the training of context and class embeddings and the generation of the prompts. All we have to supervise model training are the class labels. So we explore the feasibility and efficiency of adopting this weakly supervised learning approach in the low-resource medicine field in this work.

\subsection{Zero-shot learning and few-shot learning}
Unlike the recognition pattern of traditional machine learning methods, zero-shot learning enables the AI model to mimic human reasoning by only training on base classes and then directly recognizing new classes that have never been seen before \cite{29,30,31}. Zero-shot learning often requires auxiliary information (such as attributes of objects or related descriptions of objects, and in our research, text prompts describing pathological symptoms) to learn the semantic space. However, due to the ``seen class bias" problem \cite{30}, zero-shot learning often fails to achieve satisfactory results. Few-shot learning provides another sample-efficient learning way. By utilizing prior knowledge, few-shot learning can quickly generalize to new datasets or tasks by only learning from a small number of samples \cite{32}, which is more similar to the learning mode of human beings. Zero-shot learning and few-shot learning are suitable for the low-resource medical domain where the cost and threshold of manual labeling are extremely high. In this study, we explore the model's zero-shot and few-shot learning performance.

\section{Method}
\subsection{Overall architecture}

\begin{figure*}[!t]
	\centerline{\includegraphics[width=0.8\textwidth]{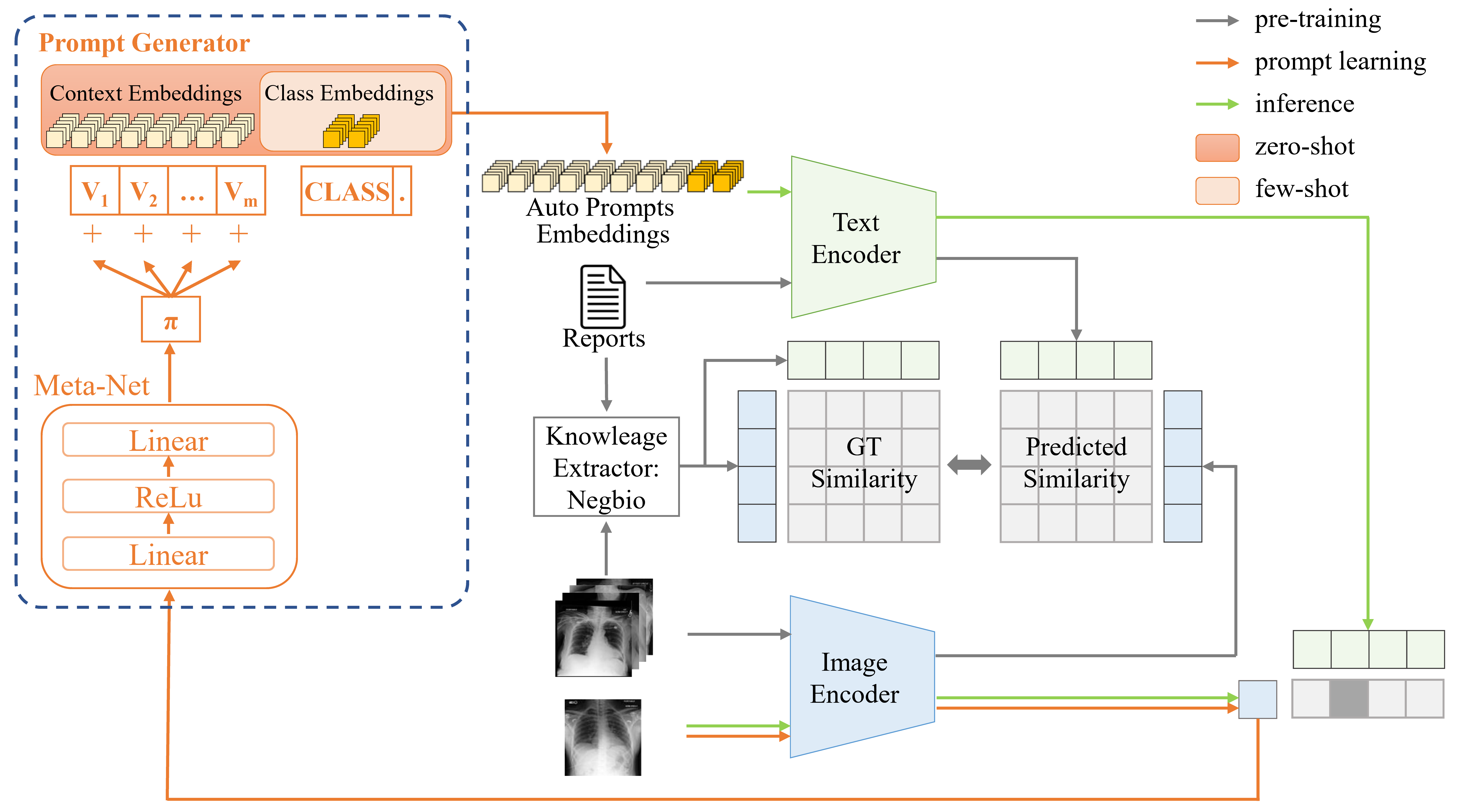}}
	\caption{The overall architecture of our proposed model $MedPrompt$. The model mainly includes pre-training (dark gray line) and prompt learning stage (orange line). In the pre-training phase, the model learns transferable representations by matching the similarity between the paired images and the reports. In the prompt learning stage, the model trains an instance-adaptive prompt generator with the help of the image embeddings and finally learns automatic prompts embeddings for zero-shot image classification. When performing few-shot image classification, the model fine-tunes the class embeddings of the prompt generator using very few images in the unseen categories, with the rest of the prompt generator remaining fixed. The green line represents zero-shot or few-shot inference on downstream image classification with the pre-trained model. Best viewed in color.}
	\label{Fig_2_architecture}
\end{figure*}

The overall architecture of our proposed model $MedPrompt$ is shown in Figure \ref{Fig_2_architecture}. The model mainly includes two training stages: pre-training (dark gray line) and prompt learning (orange line). In the pre-training stage, a total of 600,526 X-ray images and 201,063 reports from the CheXpert and MIMIC-CXR datasets are extracted by a knowledge extractor (e.g., Negbio \cite{34}) to obtain a ground-truth (GT) similarity. At the same time, these images and reports are fed into an image encoder and a text encoder, respectively, for embeddings extraction. Then, the GT similarity is used to supervise the learning of the predicted similarity between these image embeddings and text embeddings. After such pre-training, the model's text encoder and image encoder learn transferable representations, which can then be transferred to the downstream image classification task (green line). In the prompt learning stage, the model trains an instance-adaptive prompt generator with the help of the image embeddings output from the pre-trained image encoder. The prompt generator ultimately learns context embeddings and class embeddings to form automatic prompts, which can help the pre-trained model in zero-shot image classification on images of completely unseen categories. When performing few-shot image classification, the model fine-tunes the class embeddings of the prompt generator using very few images in the unseen categories, with the rest of the prompt generator remaining fixed.

\subsection{Pre-training}
We employ images and reports from two large-scale medical imaging report datasets CheXpert and MIMIC-CXR, to pre-train a large-scale vision-language model. Following MedCLIP \cite{7}, we decouple each pair of images and their reports. Instead of taking the pairing of images and their reports as the pre-training objective, we measure the similarity of each image and each report in the datasets and use similarity matching as the pre-training objective. Specifically, we focus on the 14 labeled observations included in the CheXpert dataset: No Finding, Enlarged Cardiomediastinum, Cardiomegaly, Lung Opacity, Lung Lesion, Edema, Consolidation, Pneumonia, Atelectasis, Pneumothorax, Pleural Effusion, Pleural Other, Fracture, and Support Devices. For text, we leverage the Negbio \cite{34} tool as the knowledge extractor to extract medical entities defined in the Unified Medical Language System \cite{35} from the reports to construct text GT labels $T_{GT}$ about the above 14 categories —— in the form of multi-hot vectors; for images, we use the labels extracted from their corresponding reports by the Negbio tool and method mentioned above as image GT labels $I_{GT}$. Subsequently, the cosine similarity of labels between each image and each text is calculated as the ground truth to guide the training of the pre-trained vision-language model.

\textbf{Image encoder and text encoder.} The original image $I$ and report $T$ are encoded into image embeddings and text embeddings by an image encoder $E_I$ (e.g., ResNet or ViT) and a text encoder $E_T$ (e.g., Transformer), respectively. A projection operation $proj(.)$ is used for dimensional alignment to facilitate subsequent semantic similarity calculation. Thus, the image embeddings $I_e$ and text embeddings $T_e$ can be obtained by:
\begin{equation}
	I_e=proj(E_I(I))
\end{equation}

\begin{equation}
	T_e=proj(E_T(T))
\end{equation}

\textbf{Semantic similarity matching.} The cosine similarity between image label $I_{GT}$ and text label $T_{GT}$ in the pre-trained datasets will be used as the ground-truth semantic $S$:

\begin{equation}
	S=\frac{I_{GT}\cdot T_{GT}}{\left \|I_{GT}  \right \| \left \| T_{GT} \right \| }
\end{equation}

By calculating the similarity, for an image $i$, we can obtain a normalized similarity set, where the similarity between image $i$ and text $j$ can be represented as $y_{ij}$:

\begin{equation}
	y_{ij}=\frac{\exp (S_{ij}) }{\sum_{j=1}^{N_{text}}\exp (S_{ij})}
\end{equation}

\noindent , where $S_{ij}$ is the ground-truth semantic between image $i$ and text $j$, and $N_{text}$ is the total number of text.

Similarly, the similarity $\hat{S}$ between image embeddings $I_{e}$ and text embeddings $T_{e}$ can be calculated using the following formula:

\begin{equation}
	\hat{S}=\frac{I_{e}\cdot T_{e}}{\left \|I_{e}  \right \| \left \| T_{e} \right \|  }
\end{equation}

And the predicted similarity between image $i$ and text $j$ can be represented as $\hat{y_{ij}}$:

\begin{equation}
	\hat{y_{ij}}=\frac{\exp (\hat{S_{ij}}/\tau) }{\sum_{j=1}^{N_{text}}\exp (\hat{S_{ij}}/\tau) }
\end{equation}

\noindent, where $\hat{S_{ij}}$ is the similarity between image embeddings of image $i$ and text embeddings of text $j$, $N_{text}$ is the total number of text, and $\tau$ is a learnable temperature parameter initialized at 0.07.

\textbf{Loss.} We aim to minimize the semantic loss between image and text. The semantic loss is calculated by the cross entropy loss, including an image-to-text semantic loss and a text-to-image semantic loss, which are averaged:

\begin{equation}
	L=-\frac{1}{2}
	(\frac{1}{N_{img}}\sum_{i=1}^{N_{img}}\sum_{j=1}^{N_{text}}y_{ij}log\hat{y_{ij}}+\frac{1}{N_{text}}\sum_{j=1}^{N_{text}}\sum_{i=1}^{N_{img}}y_{ji}log\hat{y_{ji}})
\end{equation}

\noindent, where $N_{img}$ and $N_{text}$ are the total numbers of images and text, respectively.

\subsection{Prompt learning}

After conducting the above pre-training, we fix the parameters of the text encoder and image encoder and proceed with prompt learning, aiming to learn a prompt generator that can generate automatic prompts embeddings for each image.

The automatic prompts generated by the model consist of several learnable words and one learnable class, finally presented as embeddings. Inspired by CoCoOp \cite{26}, the prompt generator is trained to generate prompts based on specific instances rather than a general class to learn richer transferable representations that can generalize better on unseen classes. Thus, the prompt generator consists of a learnable Meta-Net and the projection operations of context embeddings and class embeddings. Specifically, as shown in Figure \ref{Fig_2_architecture}, the features extracted by the image encoder for each image instance are processed through a Meta-Net (a two-layer bottleneck structure with Linear-ReLU-Linear) to obtain a unique conditional token $\pi $ for that instance. The conditional token is then combined with each context vector to form the final context vectors.

Therefore, for an image $i$ and a class $k$, the automatic prompt generated by the model can be denoted as:

\begin{equation}
	p_{ik}^{auto} = \left \{MetaNet(proj(E_{I}(i)) + V, C_{k}\right \}
\end{equation}

\noindent, where $V=\left \{v_1,v_2,…,v_m\right \}$ are $m$ word embeddings, each with a dimension of 512, which are randomly initialized, $m$ is the number of context tokens in the prompt, which is set to 16 in this study, and $C_{k}$ is class embeddings of class $k$.

Then, the generated automatic prompt embeddings are fed to the text encoder of the model, and the similarity $S^{auto}_{ik}$ between the generated new text embeddings $proj(E_T(p_{ik}^{auto}))$ and the image embeddings $proj(E_I(i))$ of image $i$ can be represented as:

\begin{equation}
	S^{auto}_{ik} =
	\frac{proj(E_T(p_{ik}^{auto}))\cdot proj(E_I(i))}{\left \|proj(E_T(p_{ik}^{auto})) \right \| \left \|proj(E_I(i))\right \|}
\end{equation}

Therefore, the probability of image $i$ being predicted as class $k$ can be expressed as:

\begin{equation}
	\hat{y_{ik}}=\frac{\exp (S_{ik}^{auto}/\tau) }{\sum_{k=1}^{N_{class}}\exp (S_{ik}^{auto}/\tau) }
\end{equation}

\noindent, where $N_{class}$ is the total number of classes, and $\tau$ is a learnable temperature parameter initialized at 0.07.

Then, the cross-entropy loss was calculated between the prediction probabilities of $N_{class}$ classes and the GT class of each image using formula (7).

Prompt learning is conducted before zero-shot inference and during few-shot learning. The prompt generator is trained on base classes before zero-shot classification and then performs zero-shot inference directly on unseen classes. In this process, parameters of the pre-trained model's text encoder and image encoder are fixed, while Meta-Net and the context and class embeddings corresponding to $v_1,v_2,…,v_m$ and $class$ are trained. When performing few-shot learning, parameters of the text encoder and image encoder of the pre-trained model are also fixed, and Meta-Net and the context and class embeddings corresponding to $v_1,v_2,…,v_m$ and $class$ are also trained on the base classes. Then, a few (e.g., 1,2,4,8,16) samples from unseen classes are used to fine-tune the model. Note that only the class embedding is fine-tuned at this time.

It is worth mentioning that in the process of prompt learning, for the automatic prompts to be generated, we do not have exact natural language or semantic labels that can be used to train the context and class embeddings. All we have to supervise model training are the class labels. This situation is precisely where the low-resource problem in the medical field lies —— due to the high cost and threshold of manual labeling, there are too few available exact labels. However, in the absence of sufficient exact labels, our weakly supervised learning method, which only uses class labels for training, has achieved good results, generating automatic prompts comparable to or surpassing manual prompts. This weakly supervised learning approach effectively alleviates the problem of expensive and limited manual labeling and excessive reliance on domain experts in the low-resource medical field.

\section{Experiments}

In this section, we evaluated $MedPrompt$ on four different benchmark datasets on a downstream task of medical image classification. We validated the model's performance on zero-shot inference and few-shot learning and performed ablation studies to analyze the contribution of critical factors of $MedPrompt$. In addition, we visualized and analyzed the generated prompts and discussed the limitations of this study.

\subsection{Datasets}

\textbf{CheXpert} \cite{36} is a dataset containing 224,316 chest X-ray images, covering 65,240 patients who underwent radiologic examinations at Stanford Medical Center. The dataset includes training, validation, and test sets, containing 14 observed labels. Among them, the training set includes three sets of labels automatically extracted from relevant radiology reports using various automatic labelers. Certified radiologists provide labels in validation and test sets. The dataset is of great significance for the automatic analysis and diagnosis of chest X-rays. In our study, the entire dataset participated in the unsupervised pre-training phase. To evaluate the performance of the model on zero-shot inference and few-shot learning, we followed Huang et al. \cite{6} to sample a multi-class subset \cite{7}, namely CheXpert 5$\times$200 (abbreviated as CheXpert), which contains five categories: Atelectasis, Cardiomegaly, Consolidation, Edema and Pleural Effusion, each with 200 positive samples.

\textbf{MIMIC-CXR} \cite{39} is a large medical imaging dataset with 377,111 chest X-ray images, 201,063 corresponding radiological reports, and 14 observed labels. The dataset aims to support the research of image understanding, natural language processing, and decision support in the medical domain. Similarly, the entire dataset was used for pre-training. In addition, for evaluation, we also sampled a MIMIC-CXR-5$\times$200 subset (abbreviated as MIMIC-CXR) in the same way as CheXpert-5$\times$200, with the same five categories and 200 positive samples for each category.

\textbf{COVID} \cite{40} is a publicly available X-ray image dataset on COVID-19 released by Rahman et al. in 2021. This dataset contains COVID and non-COVID labels, with a positive-to-negative sample ratio of approximately 1:1. This dataset did not participate in pre-training but only in prompt learning and evaluation. The training and test sets had 2,162 and 3,000 images, respectively.

\textbf{RSNA} \cite{41} is a chest X-ray image dataset publicly provided by the National Institutes of Health in the United States, which consists of pneumonia and normal samples. Following Wang et al. \cite{7}, we extracted a balanced subset with a 1:1 ratio of positive and negative samples. This dataset also did not participate in pre-training but only in prompt learning and evaluation. The training and test sets had 8,486 and 3,538 images, respectively.

\subsection{Baselines}

We compared our model with previous studies, including CLIP \cite{4}, ConVIRT \cite{5}, GLoRIA \cite{6} and MedCLIP \cite{7}.

\textbf{CLIP} is a pre-trained model for matching images and text released by OpenAI in early 2021. It is pre-trained directly on 4 million image-text pairs collected from the Internet and achieves state-of-the-art performance on many tasks on natural image datasets.

\textbf{ConVIRT} is dedicated to visual-text contrastive learning in medicine, learning medical visual representations directly from naturally paired medical image and text data. It is the pioneering work of multimodal contrastive learning in the medical field.

\textbf{GLoRIA} is also committed to visual-text contrastive learning in medicine and uses attention mechanisms to learn the global-local representation of images by matching words and image subregions in radiology reports. For the first time, it realizes the prompt-based zero-shot prediction for medical images after image-text pre-training.

\textbf{MedCLIP} employs an unsupervised contrastive learning approach on unpaired medical image-text data for large-scale visual-language pre-training. It utilizes BioClinicalBERT \cite{42} as text encoder and ViT as image encoder to construct the model and employs soft semantic loss to supervise the model's training. With the help of manually designed prompts, the model achieves state-of-the-art performance on downstream medical tasks such as zero-shot image classification.

\subsection{Implementation details}

We adopted the pre-trained text transformer from CLIP \cite{4} as our text encoder. For the vision encoder, we compared the CNN-based ResNet-50 \cite{43} and the transformer-based Swin Transformer \cite{44}, and finally adopted Swin Transformer as the vision encoder.

We used the CheXpert and MIMIC-CXR datasets for pre-training. We split their text data into sentences and removed all sentences with lengths under 4.

In the prompt learning stage, we held out five classes, Atelectasis, Cardiomegaly, Consolidation, Edema, and Pleural Effusion in CheXpert and MIMIC-CXR as unseen classes (to validate the zero-shot and few-shot performance of the models), and used the remaining nine classes as base classes to train the prompt generator.

We performed the following data augmentation on the images: scaling the original image to 256$\times$256, applying a random crop with a size of 224$\times$224, horizontal flipping with a probability of 0.5, color jittering the brightness of the image to a random value within [80\%, 120\%], and random affine transformation with degree sampled from [-10, 10].

We set the learning rate as 5e-5 with a learning rate warmup ratio of 0.1, weight decay 1e-4, batch size 400, and trained the model for 20 epochs. A single half-precision pre-training on 4 V100 GPUs took about 10 hours.

\subsection{Comparison to state-of-the-art methods}

\subsubsection{Zero-shot inference}

After training the prompt generator with the base classes, the model directly performs zero-shot inference on classes it has never seen before. Specifically, for CheXpert and MIMIC-CXR, there are five unseen categories: Atelectasis, Cardiomegaly, Consolidation, Edema, and Pleural Effusion. For COVID and RSNA, both the two categories in the datasets are completely new to the model. The process of performing zero-shot classification on an image is as follows: Based on the image embeddings obtained by the image encoder, the trained prompt generator generates an automatic prompt for each predicted image category. After being processed by the text encoder, these automatic prompts are turned into text embeddings of the image about each class. Then, the similarities between these text embeddings and the image embeddings of the image are calculated. The class corresponding to the text embeddings with the highest similarity is the class of the image predicted by the model.

We compared the zero-shot classification performance of our automatic prompts-based model with the previous models based on manually designed prompts, as shown in Table \ref{Table_1_zero_shot}. As can be seen from the table, our model achieves better results than all previous models on CheXpert, MIMIC-CXR, and COVID, but performs slightly worse than MedCLIP on the RSNA dataset. We speculate that the reason for the less satisfactory results on RSNA is that the labels of the base classes of CheXpert and MIMIC-CXR used to train the prompt generator are annotated by the automated tool Negbio\cite{34}. In contrast, RSNA uses manual annotations, which may have gaps with Negbio’s annotations. It is worth mentioning that the model has never seen any sample in COVID and RSNA during the large-scale image-text pre-training and prompt learning stages, but still performs well on these two datasets. Overall, it can be concluded that the prompts automatically generated by our model make a very significant contribution in assisting zero-shot image classification compared to manual prompts, resulting in an overall better generalization ability of the model.

\begin{table}[h] \footnotesize 
	\centering
	
	\begin{tabular}{*{5}{c}}
		\toprule
		Method & CheXpert & MIMIC-CXR & COVID & RSNA    \\
		\midrule
		CLIP \cite{4} & 0.2036 & 0.2254 & 0.5090 & 0.5055  \\
		ConVIRT \cite{5} & 0.4224 & 0.4010 & 0.6647 & 0.4647  \\
		GLoRIA \cite{6} & 0.4210 & 0.3382 & 0.5702 & 0.4752  \\
		MedCLIP-ViT \cite{7} & 0.5942 & 0.5024 & 0.7943 & \textbf{0.7682}  \\
		\midrule
		MedPrompt-ViT (Ours) & \textbf{0.6220} & \textbf{0.5720} & \textbf{0.7997} & 0.7284    \\
		\bottomrule
	\end{tabular}
	\caption{Performance of zero-shot image classification on four datasets. For models with manually designed prompts, we only report the results with prompt ensemble. Best performance are in bold.}
	\label{Table_1_zero_shot}
\end{table}

To further analyze the performance of zero-shot classification, we explored the classification performance of its fully supervised counterpart, which is trained with all examples on the training set. As shown in Table \ref{Table_2_full_shot}, full supervision is naturally better than direct zero-shot inference, which is undoubtedly reflected in the experimental results. Our model significantly outperforms the existing models on all four datasets, with the most significant improvement (16\%) achieved on the COVID dataset. In addition, by comparing the results of our model in Table \ref{Table_1_zero_shot} with those of the existing models in Table \ref{Table_2_full_shot}, we are pleasantly surprised to find that our model's performance of zero-shot classification without any samples for fine-tuning has surpassed the fully supervised results of the existing models on CheXpert, MIMIC-CXR, and COVID, which reflects that the model has learned transferable representations and has strong generalization ability.

\begin{table}[h] \footnotesize 
	\centering
	
	\begin{tabular}{*{5}{c}}
		\toprule
		Method & CheXpert & MIMIC-CXR & COVID & RSNA    \\
		\midrule
		CLIP \cite{4} & 0.3020 & 0.2780 & 0.5866 & 0.7303  \\
		ConVIRT \cite{5} & 0.4770 & 0.4040 & 0.6983 & 0.7846  \\
		GLoRIA \cite{6} & 0.5370 & 0.3590 & 0.7623 & 0.7981  \\
		MedCLIP-ViT \cite{7} & 0.5960 & 0.5650 & 0.7890 & 0.8075  \\
		\midrule
		MedPrompt-ViT (Ours) & \textbf{0.6580} & \textbf{0.6160} & \textbf{0.9553} & \textbf{0.8304}    \\
		\bottomrule
	\end{tabular}
	\caption{Performance of fully supervised image classification on four datasets. Best performance are in bold.}
	\label{Table_2_full_shot}
\end{table}

In addition, we compared the accuracy of zero-shot classification of different manual prompts and automatic prompts generated by our model on all four datasets, as shown in Figure 1 in Part A in the supplementary material. It can be seen that different manual prompts can lead to highly different zero-shot classification performances, which means that the quality of manual prompts has a direct and significant impact on model performance. This further indicates that the design of manual prompts depends heavily on domain experts. The method we proposed to automatically generate prompts can alleviate this dependency.

\subsubsection{Few-shot learning}

In the few-shot learning phase, the pre-trained image encoder and text encoder are fixed, and the whole prompt generator has been trained with the base classes. Then, we followed the few-shot evaluation protocol in CLIP by taking a few samples (e.g., 1,2,4,8,16) from the training set of unseen classes to fine-tune the class embeddings in the prompt generator. The model is deployed on the test set, and the results are shown in Table \ref{Table_3_few_shot}. It can be seen from the table that as the sample size used for few-shot learning gradually increases, the performance of our model on the four datasets gradually improves.

\begin{table}[h] \footnotesize 
	\centering
	
	\begin{tabular}{*{5}{c}}
		\toprule
		Method & CheXpert & MIMIC-CXR & COVID & RSNA     \\
		\midrule
		MedPrompt-ViT 0-shot & 0.6220 & 0.5720 & 0.7997 & 0.7284    \\
		MedPrompt-ViT 1-shot & 0.6315 & 0.5895 & 0.8020 & 0.7538        \\
		MedPrompt-ViT 2-shot & 0.6360 & 0.5875 & 0.8290 & 0.7665          \\
		MedPrompt-ViT 4-shot & 0.6400 & 0.5870 & 0.8627 & 0.7761          \\
		MedPrompt-ViT 8-shot & 0.6320 & 0.5815 & 0.8693 & 0.7778          \\
		MedPrompt-ViT 16-shot & 0.6500 & 0.6000 & 0.8700 & 0.8013          \\
		MedPrompt-ViT full-shot & \textbf{0.6580} & \textbf{0.6160} & \textbf{0.9553} & \textbf{0.8304} \\
		\bottomrule
	\end{tabular}
	\caption{Performance of few-shot learning of our model on four datasets. Best performance are in bold.}
	\label{Table_3_few_shot}
\end{table}

Comparing the results in Table \ref{Table_3_few_shot} with those of the SOTA model MedCLIP-ViT \cite{7} in Table \ref{Table_1_zero_shot}, it can be seen that with at most only four additional samples for fine-tuning, the classification performance of our automatic prompts model outperforms the zero-shot classification performance of SOTA manual prompts model on all datasets. It is worth noting that such a small sample cost and training cost is nothing compared to the cost of time-consuming and expert-dependent manual prompt design.

Comparing the results in Table \ref{Table_3_few_shot} with those of the SOTA model MedCLIP-ViT in Table \ref{Table_2_full_shot}, it can be seen that on the RSNA dataset, the performance of our model on 16-shot is close to the fully supervised performance of the SOTA model MedCLIP-ViT (while on the other three datasets, our model's zero-shot performance has already exceeded the fully supervised performance of all previous models, as mentioned in the zero-shot inference subsection). This is inspiring news in the field of low-resource medical data, as it means that even without sufficient available and reliable expert annotations, we can still perform exceptionally well on downstream tasks with the help of large-scale pre-training and automatically generated prompts.

Overall, the performance of prompts generated by our proposed prompt generator is better than the hand-crafted prompts, both in zero-shot inference and few-shot learning. We analyze that there are two main possible reasons for this: first, the performance of the manually designed prompts depends greatly on the domain expertise of the designing clinicians, which leads to a very unstable overall performance; second, the manually designed prompts may be too specific, which leads to the model's poor performance in the face of new samples, whereas our automatically generated prompts improve the generalization ability of the model through few-shot learning, which enables the model to learn more general patterns and thus better adapt to new inputs, including different data distributions and different specific data categories.

\subsection{Ablation studies}

\subsubsection{CNN-based architecture vs. transformer-based architecture}

\begin{figure*}[!t]
	\centerline{\includegraphics[width=0.7\textwidth]{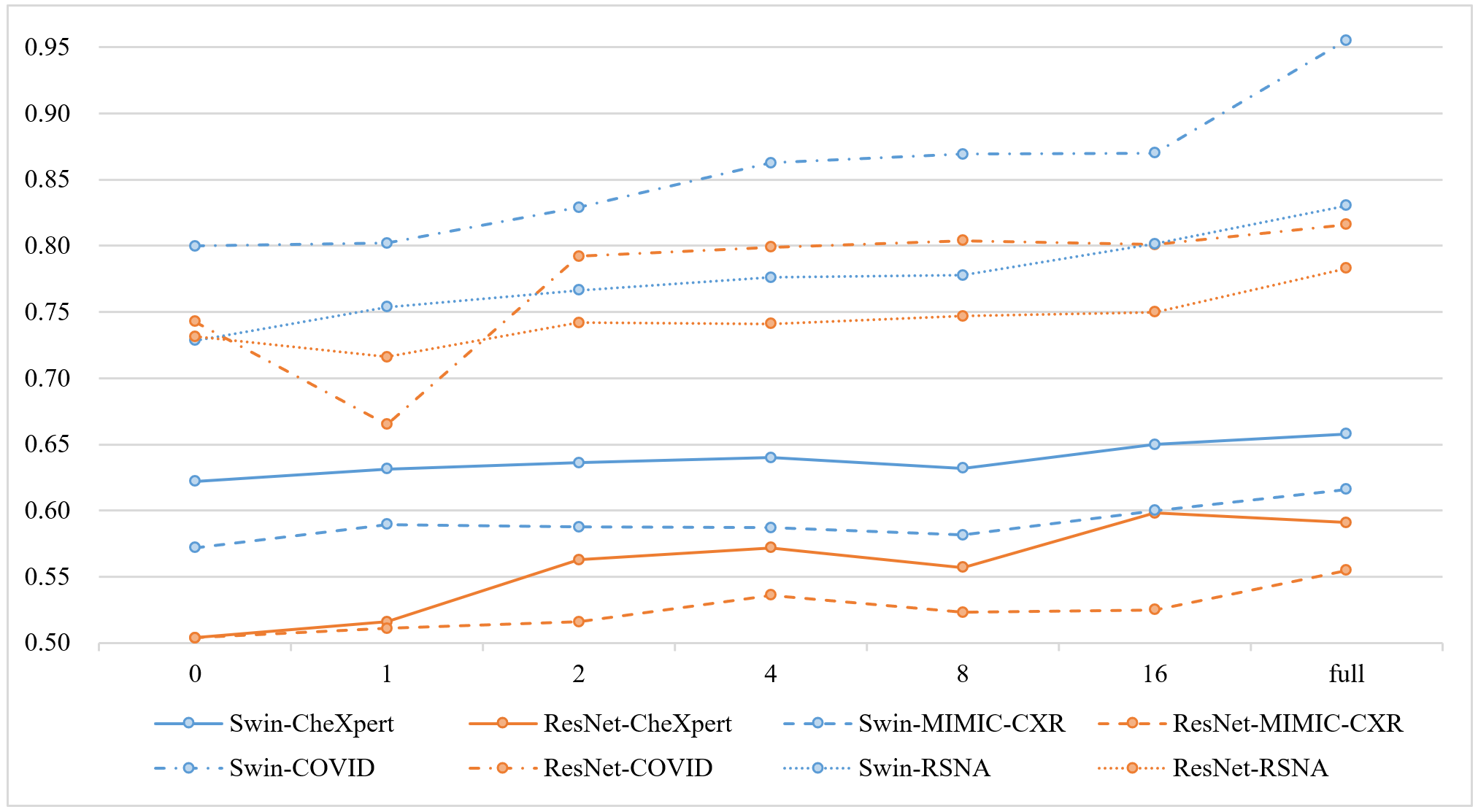}}
	\caption{Performance comparison of our model adopting the traditional CNN-based architecture ResNet-50 (the orange lines) and the Transformer-based architecture Swin Transformer (the blue lines) as the image encoder on four datasets. It can be seen that the models with ResNet-50 as the image encoder are overall inferior to their Swin Transformer counterparts. Best viewed in color.}
	\label{Fig_3_image_encoder}
\end{figure*}

We also explored the effects of adopting the classical CNN-based architecture ResNet-50 as the image encoder. As shown in Figure \ref{Fig_3_image_encoder}, the effect of using ResNet-50 as an image encoder is inferior to its Swin Transformer counterparts overall. We speculate that the main reasons are that, on the one hand, the transformer architecture has a good adaptation ability to big data, and its attention mechanism enables the model to learn the relationship between features thoroughly. On the other hand, the connection between the shifted windows in the next layer and the windows in the previous layer in the Swin Transformer significantly enhances the modeling ability, and Swin Transformer's hierarchical architecture is more conducive to capturing the visual features of different lesions with multiple scales.

\subsubsection{The effect of Meta-Net and context embeddings}

In addition, we explored the contribution of each component of the prompt generator. The learnable components of our prompt generator include Meta-Net, context embeddings, and class embeddings. Since the downstream task is image classification, the class embeddings need to be learned in any case. We validated the effectiveness of learning only class embeddings, learning only context embeddings and class embeddings, learning only Meta-Net and class embeddings, and learning a complete prompt generator, respectively. Note that the above component learning only refers to the training with base classes before few-shot learning. In any case, only class embeddings will be learned when performing few-shot learning on unseen classes. Table \ref{Table_4_ablation} presents the average results of zero-shot inference, few-shot learning, and full supervision on these four ablation versions over four datasets. We can conclude from the table that both the Meta-Net and the context embeddings contribute significantly to the performance improvement of the model. When the two are combined into the prompt generator, the model's performance has made a great leap, with zero-shot performance improving by 3\% and full supervision performance improving by 4.6\%. We speculate that the main reason is that MetaNet has learned richer transferable representations from each specific instance, which helps to better generalize the model to unseen categories. Combining these representations and context embeddings makes the generated automatic prompts more adaptable to instances in previously unseen classes.

\begin{table*}[h] \footnotesize 
	\centering
	
	\begin{tabular}{*{10}{c}}
		\toprule
		\multirow{2}{*}{Meta-Net} & Context  & Class  & \multirow{2}{*}{0} & \multirow{2}{*}{1} & \multirow{2}{*}{2} & \multirow{2}{*}{4} & \multirow{2}{*}{8} & \multirow{2}{*}{16}  & \multirow{2}{*}{full}    \\
		& Embeddings & Embeddings &  &  &  &  &  &  &      \\
		\midrule
		&  & $\surd$ & 0.6508 & 0.6877 & 0.6894 & 0.6886 & 0.6936 & 0.6994 & 0.7190 \\
		& $\surd$ & $\surd$ &  0.6566 & 0.6738 & 0.6810 & 0.6943 & 0.7011 & 0.7021 &  0.7158 \\
		$\surd$  &  & $\surd$  & 0.6675 & 0.6757 & 0.6793 & 0.6946 & 0.7005 & 0.7027 & 0.7193 \\
		$\surd$  & $\surd$  & $\surd$  & \textbf{0.6805} & \textbf{0.6942} & \textbf{0.7048} & \textbf{0.7165} & \textbf{0.7152} & \textbf{0.7303} & \textbf{0.7649}        \\
		\bottomrule
	\end{tabular}
	\caption{The average results of zero-shot learning, few-shot learning, and full supervision for these components and their combinations over four datasets. The best results are highlighted in bold.}
	\label{Table_4_ablation}
\end{table*}

In addition, we compared the training and test loss curves, accuracy curves and ROC curves of our proposed models, as well as other evaluation metrics such as precision, specificity, sensitivity and F1-Score of models using our proposed prompt generator with different learnable components, and visualize the confusion matrixes of our final model, as detailed in Part B in the Supplementary Material.

\subsection{Further analysis}

\subsubsection{What do the automatically generated prompts look like?}

We further explored what the prompts automatically generated by our model looked like. We searched within the pre-trained medical text corpus for medical-related words closest to the context embeddings learned by our model based on Euclidean distance. The top 30 closest words are presented in Table \ref{Table_5_closest_words} in ascending order of their distances. From the table, we can see that the prompts generated by the model include not only professional medical terms, such as kaif, knot, potassic, skelet, antioxidant, serine, but also general medical words, such as circulare, procedure, surround, transplant, capsule, medics, examined. It indicates that our model has learned some common knowledge to some extent when generating automatic prompts. This is very enlightening, indicating that our model may have the potential to be expanded to a general multimodal medical recognition model.

\begin{table*}[h] \footnotesize 
	\centering
	
	\begin{tabular}{{|c|c|c|c|c|c|c|c|c|}}
		%\toprule
		\hline
		$\sharp$ & word & distance & $\sharp$ & word & distance & $\sharp$ & word & distance \\
		\hline
		1 & kaif & 0.4515 & 11 & antioxidant & 0.4892 & 21 & infringe & 0.4977 \\
		\hline
		2 & knot & 0.4532 & 12 & IHRI & 0.4943 & 22 & eucalyptol & 0.4977 \\
		\hline
		3 & atility & 0.4570 & 13 & recognizable & 0.4952 & 23 & philosophy & 0.4981 \\
		\hline
		4 & potassic & 0.4653 & 14 & characteristics & 0.4953 & 24 & medics & 0.4990 \\
		\hline
		5 & disciplines & 0.4660 & 15 & born & 0.4962 & 25 & recipient & 0.5088 \\
		\hline
		6 & skelet & 0.4709 & 16 & surround & 0.4965 & 26 & serine & 0.5089 \\
		\hline
		7 & circulare & 0.4822 & 17 & ingredient & 0.4970 & 27 & jurisdictional & 0.5135 \\
		\hline
		8 & occupational & 0.4850 & 18 & transplant & 0.4970 & 28 & studies & 0.5151 \\
		\hline
		9 & procedure & 0.4883 & 19 & capsule & 0.4974 & 29 & examined & 0.5170 \\
		\hline
		10 & wearable & 0.4890 & 20 & accommodate & 0.4976 & 30 & practitioner & 0.5175 \\
		\hline
		
		%\bottomrule
	\end{tabular}
	\caption{The top 30 closest words to the context embeddings learned by our model based on Euclidean distance, sorted in ascending order of their distances.}
	\label{Table_5_closest_words}
\end{table*}

\begin{figure*}[!t]
	\centerline{\includegraphics[width=0.7\textwidth]{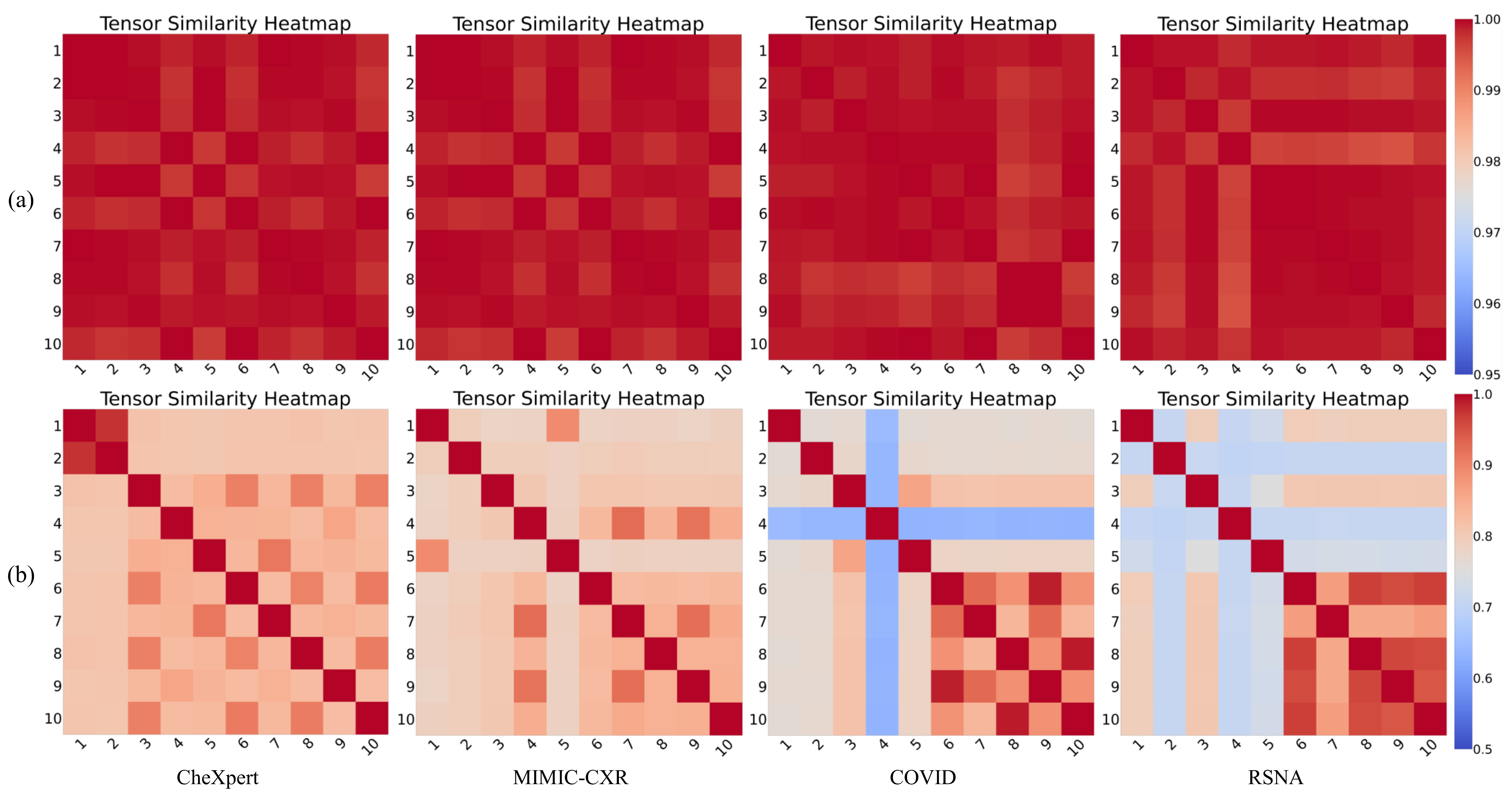}}
	\caption{Visualization of prompts similarities on four datasets. (a) The similarities of prompts automatically generated by the model. (b) The similarities of manually designed prompts. The ten images based on which the model automatically generates the prompts and the ten manual prompts are randomly selected. Best viewed in color.}
	\label{Fig_4_prompts similarities}
\end{figure*}

In addition, we randomly selected ten images in each dataset and visualized the cosine similarity of the contexts embeddings in the corresponding ten prompts automatically generated by the model, as shown in Figure \ref{Fig_4_prompts similarities} (a). We are surprised to find that the contexts generated by the model are surprisingly similar for different categories of images from different datasets. This further validates that the model has learned some common knowledge, which is consistent with the observation from the closest 30 words mentioned above.

As a comparison, we also randomly selected ten manually designed prompts for each dataset and calculated the cosine similarity between these prompts. As shown in Figure \ref{Fig_4_prompts similarities} (b), the overall similarity of the manual prompts is relatively low, indicating that the manual prompts need to be carefully designed according to the characteristics of different categories of images in different datasets. Such a process undoubtedly relies excessively on the professional knowledge of domain experts and requires a lot of effort and time.

\subsubsection{What the image encoder of our proposed model learn with the promotion of pre-training and the prompt generator?}

We generate Class Activation Maps (CAMs) for the last layer of feature maps extracted by the image encoder of our proposed model. Figure \ref{Fig_5_CAMs} shows some of the cases of different categories on the four datasets. It can be observed from these cases that our model's image encoder can efficiently learn the most noteworthy regions in an X-ray image. Taking the categories of cardiomegaly and consolidation on CheXpert for example (as shown in the second and third columns of Figure \ref{Fig_5_CAMs} (a), respectively), the model accurately identified the cardiac region and the lung region with obvious consolidation, respectively. This indicates that the model has strong representation learning ability. Also, as shown in the cases on COVID and RSNA (Figure \ref{Fig_5_CAMs} (c) and (d)), it can be seen that for images with disease symptoms, such as images of COVID and pneumonia, the model can accurately identify the corresponding lung or lesion location, while for non-COVID and non-pneumonia images, the CAMs of the model's feature maps also accurately reflect that there are no focal regions to which attentions need to be paid.

\begin{figure*}
	\centering
	\subfigure{\includegraphics[width=0.85\textwidth]{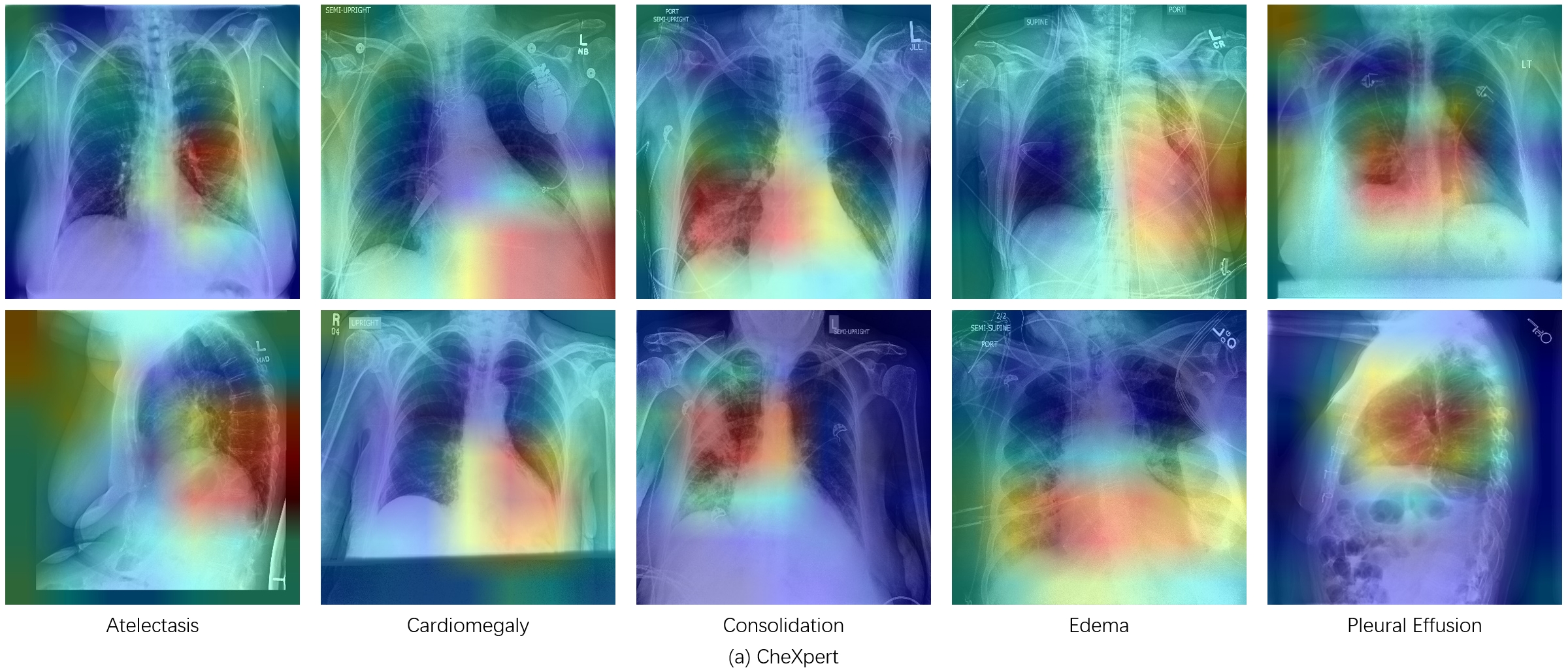}}
	\subfigure{\includegraphics[width=0.85\textwidth]{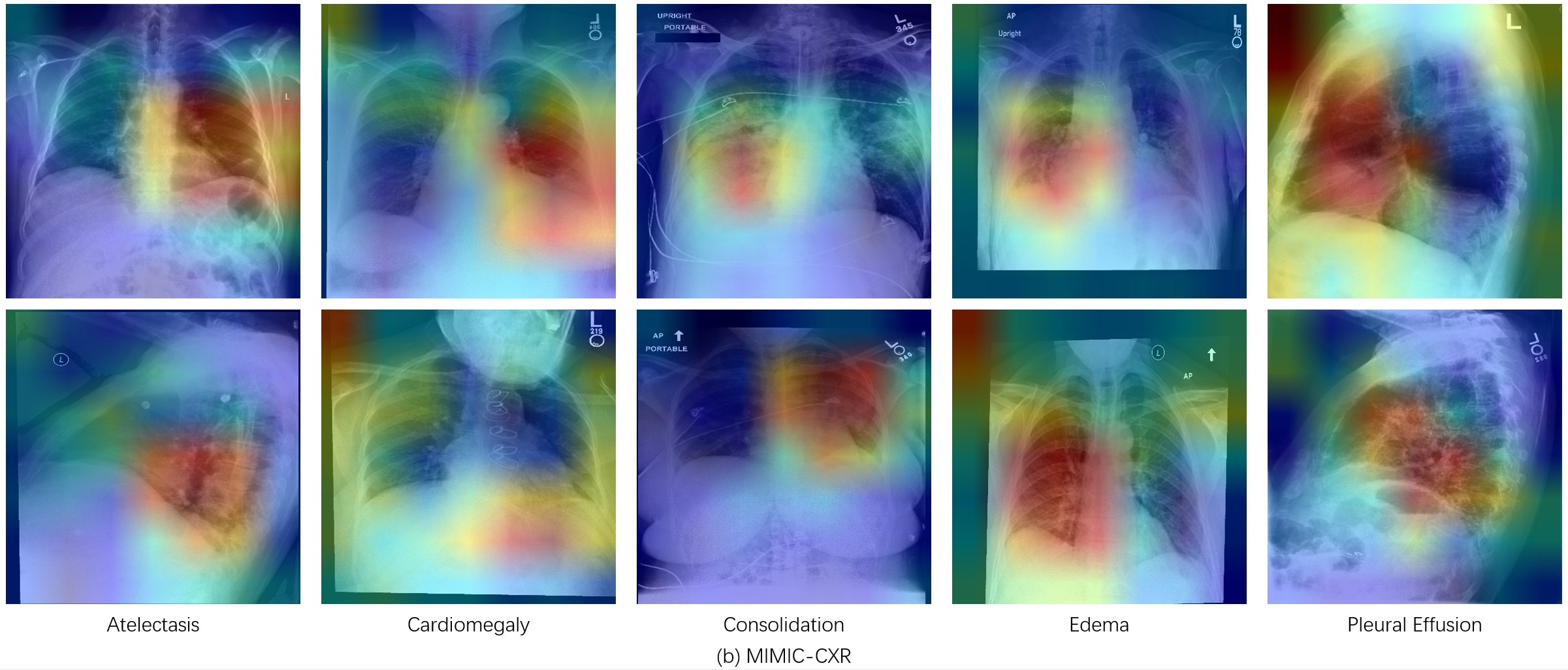}}
	\subfigure{\includegraphics[width=0.85\textwidth]{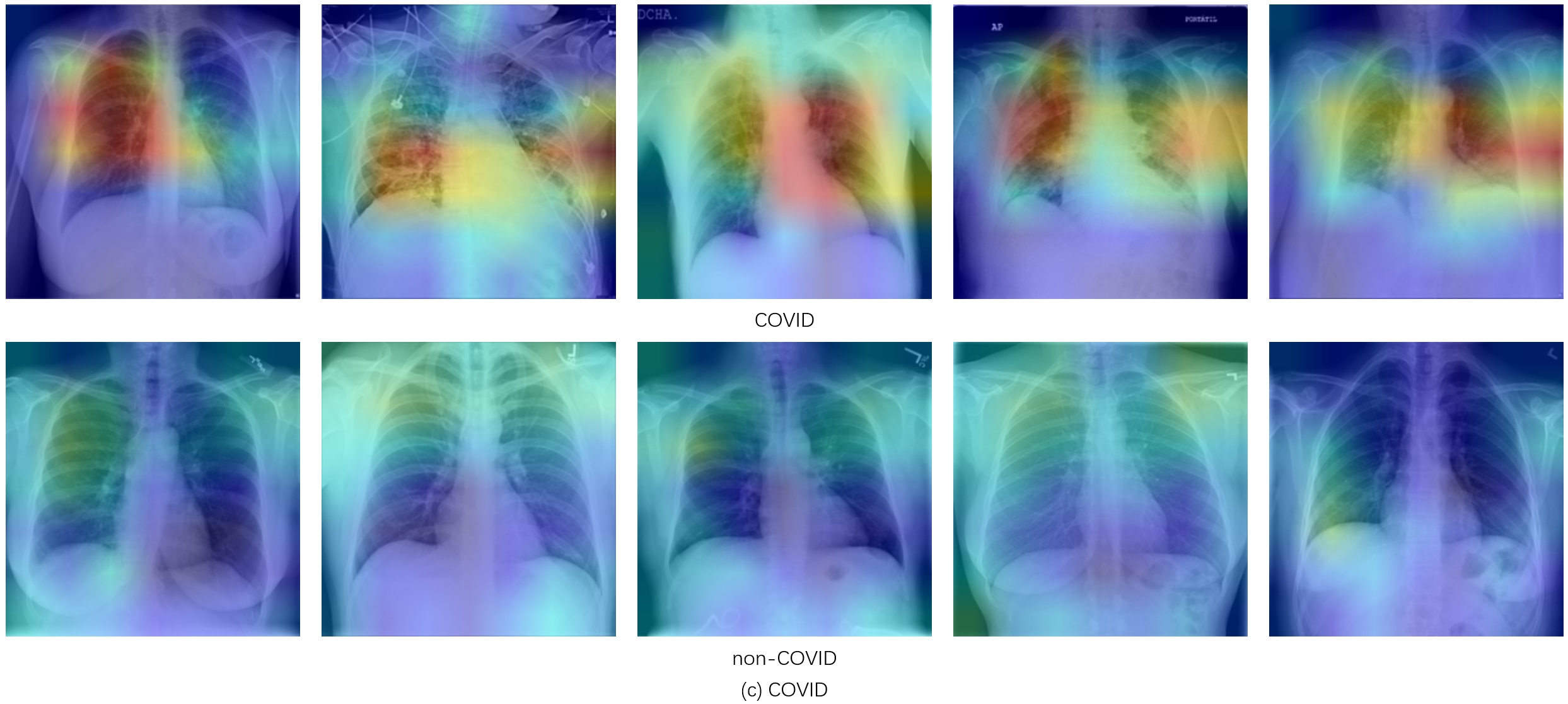}}	
\end{figure*}
\addtocounter{figure}{0}
\begin{figure*}
	\addtocounter{subfigure}{2}
	\centering
	\subfigure{\includegraphics[width=0.85\textwidth]{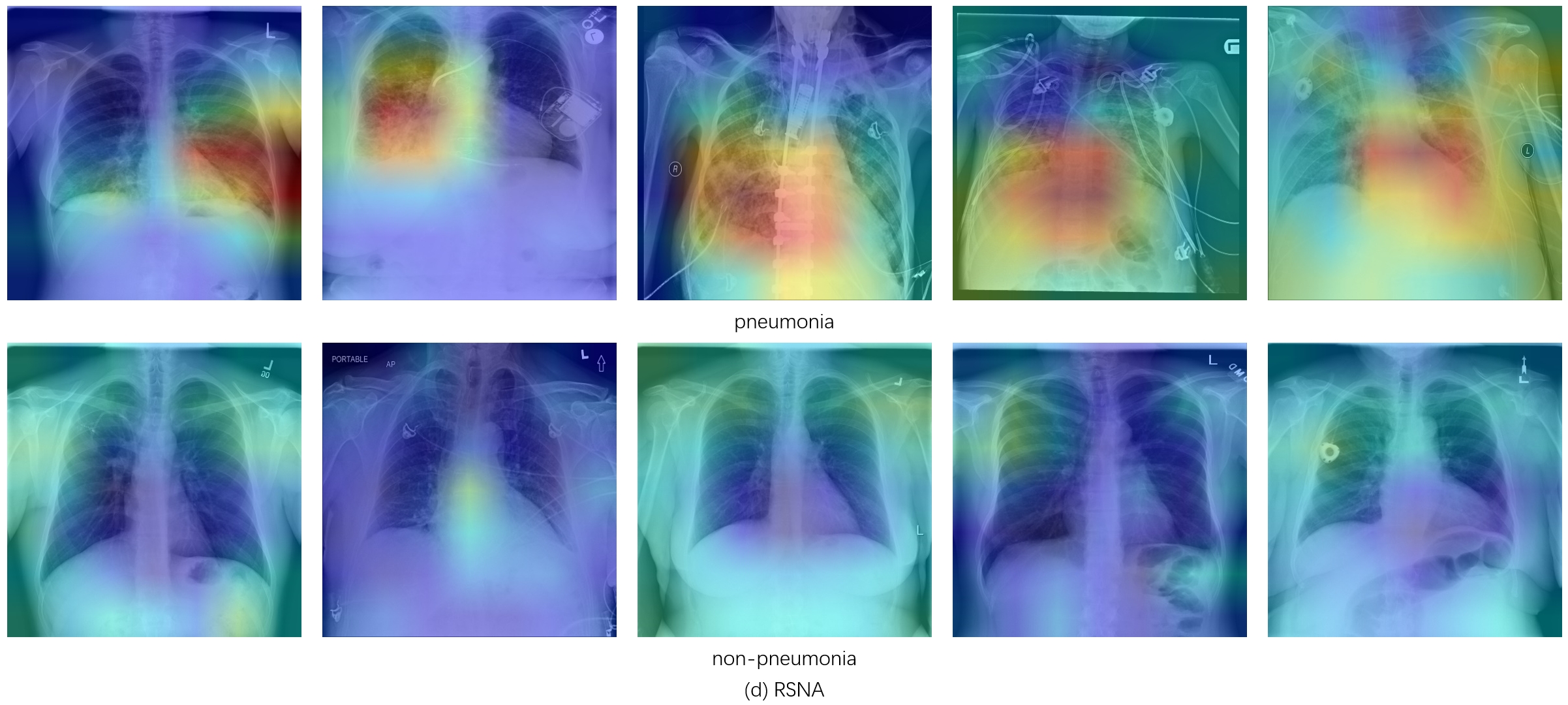}}
	\caption{Class activation maps for images of each category in the four datasets. It can be observed from these cases that our model's image encoder can efficiently learn the most noteworthy regions in an X-ray image.}
	\label{Fig_5_CAMs}
\end{figure*}

\subsubsection{Analysis of parameters and FLOPs}

\begin{figure*}[!t]
	\centerline{\includegraphics[width=0.7\textwidth]{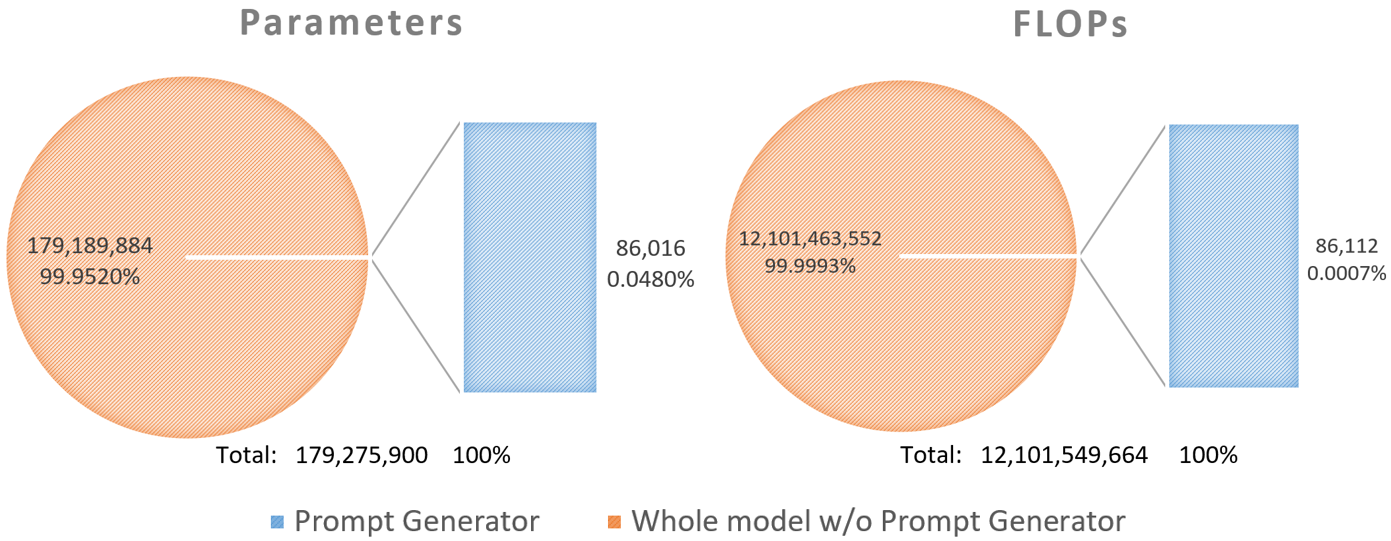}}
	\caption{Schematic diagram of the numbers of parameters and FLOPs and their propotions of prompt generator, the whole model w/o prompt generator, and the whole model.}
	\label{Fig_6_paras_and_flops}
\end{figure*}

We also analyzed the number of parameters and floating-point operations per second (FLOPs) of the prompt generator. Including Meta-Net, context embeddings, and class embeddings, the entire prompt generator has only 86,016 parameters and 86,112 FLOPs, accounting for only 0.0480\% of the 179,275,900 parameters and 0.0007\% of the 12,101,549,664 FLOPs of the entire model respectively, as shown in Figure \ref{Fig_6_paras_and_flops}. Therefore, the prompt generator is lightweight, with low extra parameters and computational overhead, and thus has the potential to be embedded into any network architecture, including large network architectures and end-side or mobile network architectures.

\subsubsection{Advantages and limitation}

To the best of our knowledge, this is the first model to automatically generate medical prompts. There are two most significant advantages of the proposed method: First, it can utilize a large amount of unlabeled medical imaging and textual data for unsupervised pre-training, obtaining a large-scale pre-trained vision-language model with strong transferable representation capabilities specifically for low-resource medical scenarios; Second, it can automatically generate high-quality medical prompts for the transfer learning of the large-scale pre-trained model based on different datasets, thereby relieving the pressure on clinicians to manually design prompts. These two advantages enable the model to get rid of the limitation of being constrained by the small amount of labeled data and reduce the great dependence on domain experts to the greatest extent. This is meaningful for advancing the exploration of intelligence and automation in the low-resource medical field. In addition, our proposed prompt generator for automatically generating medical prompts is lightweight, with only 0.08M parameters and 8e-5 FLOPs, accounting for only 0.0480\% and 0.0007\% of the entire model, respectively, which indicates that it has the potential to be embedded into any network architecture, whether it is a large network architecture, or an end-edge or mobile network architecture.

However, there are some limitations in our approach. First, the model may suffer from failure when confronted with a dataset of completely new diseases. For example, as can be seen from the experimental results in this work, the model does not perform satisfactorily in zero-shot classification on the RSNA dataset, which has never been seen during pre-training. As we analyzed in Subsection 4.4.1, this may be caused by the different annotation methods used in the pre-training datasets and the RSNA dataset. Thus, the representation transferability of the current model is still limited by the pre-trained data it has seen. A possible solution is to use more real-time updated data on the Internet for pre-training to improve the representation transferability of the model. Second, it is foreseeable that our method may not perform well in the recognition of medical images in other modalities (such as MRI, CT, etc.) that it has never seen before. This is quite intuitive since there are very significant differences between different medical imaging modalities. In future research, we will consider pre-training the large-scale vision-language model in our method on more diverse data so that it can learn the differences between different modalities, and expanding the model into one that can handle a wide variety of image modalities. Third, our method has only been validated on the medical image classification task, lacking validation on other downstream clinical tasks such as medical image retrieval, segmentation, detection, and prognosis prediction. In future studies, we will verify the transferable representation capabilities of the model in more tasks. Finally, the heterogeneity of cross-modal and cross-center data, the interpretability of the model itself, and the mixed nature of medical data in real clinical scenarios can greatly affect the performance of the model. Therefore, the robustness and generalization ability of the method remains to be fully validated before applying it to real clinical scenarios.

\section{Conclusion}

In this work, we propose a weakly supervised prompt learning method that can automatically generate medical text prompts for large-scale pre-trained vision-language models. On the one hand, the generated medical text prompts can assist in effectively transferring the pre-trained vision-language models to downstream clinical tasks such as medical image classification; On the other hand, the automatic generation of medical text prompts has freed clinicians from the pressure of manual prompt design. To the best of our knowledge, this is the first model to automatically generate medical prompts. Abundant experimental results indicate that our proposed method achieves state-of-the-art performance in the few-shot and zero-shot classification of unseen classes. Moreover, the prompt learning module is so lightweight that it has the potential to be embedded into any network architecture.

\section*{Declaration of competing interest} 

None.

\section*{Acknowledgements}
This work was supported in part by the National Natural Science Foundation of China (Grant No. U1811461), the Key Areas Research and Development Program of Guangdong (Grant No. 2018B010109006), and Guangdong Introducing Innovative and Entrepreneurial Teams Program (Grant No. 2016ZT06D211).

\end{document}